\documentclass{article}

\usepackage[dblblindworkshop,final]{neurips_2026}

\usepackage[utf8]{inputenc} 
\usepackage[T1]{fontenc}    
\usepackage{hyperref}       
\usepackage{url}            
\usepackage{booktabs}       
\usepackage{amsfonts}       
\usepackage{nicefrac}       
\usepackage{microtype}      
\usepackage{xcolor}         

\usepackage{makecell} 

\usepackage[compact]{titlesec}
\titlespacing*{\section}{0pt}{6pt}{2pt}
\titlespacing*{\subsection}{0pt}{4pt}{1pt}
\titlespacing*{\subsubsection}{0pt}{3pt}{1pt}

\newenvironment{smitemize}%
{\begin{list}{$\bullet$}%
   {\setlength{\parsep}{0pt}%
	\setlength{\topsep}{0pt}%
    \setlength{\itemsep}{0pt}
    \setlength{\leftmargin}{6mm}}}
{\end{list}}

\newcommand{\longpage}{\enlargethispage{\baselineskip}}

\title{The Implications of Linguistic Illegibility for LLM Security}
\workshoptitle{Foundations of Language Model Security}

\author{%
  James Mickens \\
  Harvard University\\
  \texttt{mickens@g.harvard.edu} \\
}

\begin{document}

\maketitle

\begin{abstract}
LLMs are trained to generate natural language.
However,
various strands of evidence indicate that
an LLM's externalized linguistic outputs
and mechanistically-extracted linguistic features
can be an unreliable lens for
understanding internal model computation.
We introduce the term \textit{linguistic illegibility}
to broadly refer to scenarios
in which an LLM's
externalized or mechanistically-probed language artifacts
fail to represent
how the model actually thinks.
We argue that the specter of linguistic illegibility
is unavoidable for LLMs whose internal computations
are not directly expressed via language,
but rather math over activation spaces
(with lossy translations between activation spaces
and natural language happening at the bookends).
If linguistic illegibility is always possible,
then security mechanisms that rely on
a model's linguistic self-reporting
(e.g., chain-of-thought monitoring,
constitutional self-critique,
activation probing for
linguistically-defined feature vectors)
can never be completely sound;
the model sandbox will always need
isolation techniques whose guarantees
do not depend on
reading a model's linguistic state at all.
We argue that observing a model's outputs
using taint tracking
is a promising approach for an effective sandbox:
regardless of how a model linguistically self-reports,
a taint tracking policy can define, a priori,
various pieces of system state that
should never be influenced by model-produced data.
We also discuss several additional sandboxing mechanisms
(e.g., robust virtualization,
third-party auditing of sandboxing configurations)
which collectively provide a critical floor
beneath linguistic monitoring,
and would have mitigated recent sandbox exploits
by frontier models.
\end{abstract}

\section{Introduction}
\label{sec:introduction}

Large language models (LLMs)
are increasingly entrusted with sensitive data,
and increasingly empowered to
autonomously interact with the physical world.
Two critical security challenges arise.
The first involves
\textbf{detection of misaligned intent:}
how can we detect when LLMs are planning
to carry out behavior with harmful consequences?
The second challenge is
\textbf{disruption of misaligned action:}
in situations where misaligned planning
cannot be detected and derailed,
how can we prevent a model
from actually engaging with the external world
in malicious ways?

In the classic literature on systems security,
the disruption of misaligned action
is called ``sandboxing.''
A sandbox ensures that
code which tries to do harm cannot.
Classic sandboxing techniques like
system call filtering~\cite{demarinis2020sysfilter,linuxseccomp},
hardware virtualization~\cite{liu2024riscv,tichon2017hardware,uhlig2005intel},
and network firewalling~\cite{hsieh2024netvigil,kubernetesnetworkpolicies,scarfone2009firewalls}
are now employed in the LLM context too.
For example,
best practice dictates that
an inference-as-a-service deployment
should be decomposed using least privilege~\cite{provos2003privilege},
such that each component
(e.g., the inference engine,
the RAG database,
and the execution environment
for model-generated code)
is placed in
a tightly-restricted container or VM~\cite{langchain2026sandbox,langchainsecuritypolicy}.

Detection of misaligned intent
is also a goal of classic systems security.
In the ideal world,
harmful code would announce its goals
using natural language that
could be analyzed in straightforward ways.
However,
traditional programs
(e.g., web browsers and video games)
plan via code and
externalize intent via
system calls, managed runtime interactions,
or other non-lingual means;
as a result,
intent detection traditionally relied
on static code analysis~\cite{kinder2005detecting,moser2007limits,schmeelk2015}
or dynamic observation of
environmental interactions~\cite{baker2025monitoring,egele2012survey,zhang2019dynamic}.

Unlike a traditional program,
an LLM uses natural language
as a primary interface with humans.
Thus,
a popular approach for detecting misaligned LLM intent
is to treat a model's own language outputs
as a diagnostic signal.
Chain-of-thought traces are
examined for signs of
reward hacking or deception~\cite{baker2025monitoring,korbak2025chain};
constitutional self-critique asks a model
to evaluate its own outputs against
principles formulated in natural language~\cite{bai2022constitutional,chaconmenke2025effective};
activation probes are trained to detect
linguistically-framed concepts such as
truthfulness, harmfulness, or
planned deception~\cite{burns2022discovering,macdiarmid2024simple,marks2024geometry}.
These approaches have an obvious intuitive appeal:
a model trained to receive natural-language inputs
and generate natural-language outputs
should have internal behaviors that
are legible via linguistic analyses.

However,
the robustness of these approaches
is tied to whether
the \textit{linguistically-decodable} aspects
of an LLM's state
actually reflect the model's true intent.
We use the phrase ``linguistically decodable''
by way of analogy to the phrase ``linearly decodable'' that is often used
in the mechanistic interpretability literature.
A linearly-decodable feature of model cognition
is one that can be detected via
a linear function applied to
a hidden layer's activations.
A linguistically-decodable feature is
one that is defined in terms of natural language,
and is extracted directly from
a model's natural-language outputs
or indirectly via mechanistic probes
like sparse autoencoders~\cite{bricken2023monosemanticity}
or logit lenses~\cite{nostalgebraist2020logitlens}
which map language-level concepts like
``happy'' or ``conflicted'' to
functions of activation vectors.

So, are an LLM's linguistically-decodable features
guaranteed to reflect a model's internal computations?
A growing body of empirical work says ``no,''
and we argue that the problem is intrinsic to
how LLMs are constructed (\S\ref{sec:ling-ill}):
there is no way to eliminate the possibility
of semantic divergence between a model's natural-language outputs
and its true planning goals.
This lingering threat is particularly alarming
given the rapid increase in model capabilities.
As those capabilities continue to improve,
we argue that the AI safety community must pay more attention
to isolation mechanisms that do not rely
the extraction of natural-language observations
from increasingly byzantine models.

In Section~\ref{sec:stronger-sandbox},
we propose some concrete design principles for
a systems-level, non-linguistic sandbox
for misaligned frontier models.
A core idea is to aggressively track how model-generated outputs
are used to derive new data and
influence critical pieces of system state.
Such taint tracking is critical because
in many cases
we can define, a priori,
system state that models should not influence
\textit{regardless of the model's ostensible or true planning process.}
For example,
if a model should be isolated from the public internet,
there is never a reason that model-influenced bytes
should appear in an outbound network packet.

\section{Linguistic Illegibility: Examples and Security Implications}
\label{sec:ling-ill}


\subsection{When Can Illegibility Occur?}
\longpage
\paragraph{Chains of thought:} Asking an LLM
to explain its thought process
leads to performance gains
in many settings~\cite{kojima2022zeroshot,wei2022chain}.
This gain is intuitively reasonable:
if LLMs internally use linguistic reasoning,
then providing the model with a ``linguistic scratchpad''
should give the model additional opportunities
to plan and self-correct.
However,
empirical work has shown that chains of thought
are not necessarily a faithful representation
of a model's true planning process.
For example:
\begin{smitemize}
  \item Turpin et al. demonstrated that if
        the chain-of-thought exemplars provided in
        system prompts contain a hidden bias
        (e.g., the correct answer to a multiple choice
        question is always ``(a)''), models
        often do not mention those biases in
        chains of thought, and instead rationalize
        away incorrect answers that were clearly
        influenced by the bias~\cite{turpin2023unfaithful}.
  \item To further test the extent to which chains of
        thought can be post-hoc rationalizations
        for model output, Lanham et al. took real
        prompt+CoT pairs generated by an LLM,
        edited the CoT by truncating it, inserting
        a logically incorrect sentence, or correctly
        paraphrasing the original CoT, and then examined the resulting model output~\cite{lanham2023measuring}.
        Model accuracy was only minimally affected by 
        paraphrased CoTs. However, truncated or incorrect 
        CoTs had different effects on different model 
        tasks, sometimes hurting accuracy but sometimes
        having little impact; for the latter kinds of
        tasks, model reasoning could not have relied
        on the linguistic content of CoT tokens.
  \item When an LLM is trained via reinforcement learning,
        the model might try to ``reward hack''
        (i.e., maximize its training reward by generating
        solutions that are misaligned with respect to
        human-desired goals, but nonetheless maximize
        the scoring function). Chen et al. found that
        Claude 3.7 Sonnet verbalized its use of discovered
        hacks less than 2\% of the time in 5 out of 6
        evaluation scenarios~\cite{anthropic2025rewardhacking}.
  \item Each token that a model generates
        is created via hidden-layer computations.
        Thus, asking a model to give a chain of thought
        before answering a question
        will give the model extra computational capacity
        regardless of whether the resulting CoT tokens
        are linguistically legible.
        As an extreme demonstration of this idea,
        Pfau et al. showed that a model whose
        chain of thought is just dots (``...'')
        can nonetheless use the associated compute
        to solve problems that are intractable for
        a model with an equivalent architecture but
        no chain of thought~\cite{pfau2024dot}.
\end{smitemize}
Other examples of
CoTs being linguistically illegible abound~\cite{arcuschin2025,arnav2025}.

\vspace{-1mm}

\paragraph{Constitutional self-critique:}
Constitution-based alignment (CAI) methods
use a natural-language document to
define what ``good model behavior'' is;
the document guides fine-tuning,
with the goal of nudging the model into
a more aligned state~\cite{bai2022constitutional}.
The basic CAI approach has two phases:
\begin{smitemize}
  \item A pretrained model $M$ receives a prompt,
        generates a response, and is requested to 
        critique the response according to alignment 
        principles contained in a natural-language 
        constitution. Examples of such principles might be
        ``Do not harm users'' or ``Always be polite.''
        The model revises the response according to the 
        constitution. The <original prompt, 
        refined response> pair is then used to
        fine-tune $M$. This process is repeated across 
        many pairs, producing the initial fine-tuned
        model $\hat{M}$.
  \item $\hat{M}$ is iteratively provided a prompt and
        asked to generate two different responses.
        A separate judge model $J$ ranks the two
        responses. The results of many such pairwise
        comparisons are used to train a reward model
        which is then used to refine $\hat{M}$ via
        reinforcement learning.
\end{smitemize}
The potential illegibility of a model's output tokens
threatens the effectiveness of CAI.
For example,
suppose that
the un-fine-tuned $M$ is a ``sleeper agent'' model
that emits harmful code
only when a trigger word
is in a prompt~\cite{hubinger2024sleeper}.
In phase one of CAI,
when $M$ is asked to
critique its response for a triggering prompt,
the model might affirm the principle
``All code that I emit should be safe''
while nonetheless emitting unsafe code.
Later,
in the second phase of CAI,
$\hat{M}$ can try to reward-hack the evaluator model $J$,
producing responses whose overt linguistic content
please the judge but do not reveal
$\hat{M}$'s true strategy for task completion~\cite{yang2025superficial}.

\vspace{-1.5mm}
\paragraph{Mechanistic interpretability:}
The goal of mechanistic interpretability
is to observe a model's internal activation vectors
and identify the semantic concepts (or ``features'')
that are relevant to the model's planning~\cite{nanda2023mechinterp}.
From the security perspective,
mechanistic interpretability is
an attractive foundation for runtime sandboxing:
if a monitor can detect that a model
is internally ruminating on unsafe features,
the model can be halted or nudged towards
less harmful features~\cite{stolfo2024,turner2023}.

A common assumption in mechanistic interpretability is
the linear representation hypothesis,
which states that a model represents features
as directions in the activation space~\cite{park2024linear}.
For example,
let $\mathbf{h_{\ell,t}}$ represent
a model's hidden-layer activations
at layer $\ell$ and forward pass $t$.
\begin{smitemize}
  \item We could train a linear probe~\cite{alain2016understanding}
        of the form $\mathbf{f} = softmax(\mathbf{Wh_{\ell,t}} + \mathbf{b})$,
        where $\mathbf{W}$ is a learned matrix,
        $\mathbf{b}$ is a learned vector,
        and $\mathbf{f}$ is a probability vector
        which captures the likelihood that a particular
        feature is represented by an activation.
        For example,
        Gurnee and Tegmark demonstrated that
        Llama-2 has linearly-decodable features
        representing geographic locations for physical
        places such as landmarks, such that 
        when prompted about a specific place,
        the longitude and latitude of that place
        are recoverable~\cite{gurnee2024space}.
  \item A sparse autoencoder maps a hidden layer
        $\mathbf{h_{\ell,t}}$ of dimension $d$ to a
        much larger feature space of dimension $f > d$; the
        encoding looks like $\mathbf{x} = g(\mathbf{W_e h_{\ell,t}} + \mathbf{b_e})$
        where $g$ is a nonlinear activation function.
        The autoencoder then maps $\mathbf{x}$ back to the 
        original dimensionality via the decoding 
        projection $\mathbf{\hat{h}_{\ell,t}} = \mathbf{W_d x} + \mathbf{b_d}$. The autoencoder
        is called ``sparse'' because the function for
        training loss not only encourages reconstruction 
        accuracy, but also rewards $\mathbf{x}$ values
        that have few non-zero entries; the hope is that 
        the resulting autoencoder will be able to generate 
        a sparse feature vector for the target model's
        $\mathbf{h_{\ell,t}}$. For example, Cunningham
        et al. built a sparse autoencoder for Pythia
        models and empirically identified various 
        dimensions in the expanded space that corresponded
        to natural-language concepts like indirect
        objects~\cite{huben2024sparse}.
  \item A logit lens~\cite{nostalgebraist2020logitlens} 
        uses a transformer's
        token vocabulary as the feature dictionary,
        and simply applies the transformer's final-layer
        unembedding matrix $\mathbf{U}$ to a
        non-final-layer activation vector 
        $\mathbf{h_{\ell,t}}$. The probability vector
        $\mathbf{Uh_{\ell,t}}$ is treated as the likelihood
        that various features (as named by tokens)
        are the focal areas of model planning at
        layer $\ell$.
\end{smitemize}
Crucially,
all of these mechanistic probes
are linguistically anchored in some way,
searching for features that
are defined in natural language.
Logit lenses literally use the token vocabulary
as the feature dictionary.
With sparse autoencoders,
the semantic meanings of dimensions
in the feature space
are determined by a second LLM
tasked to observe how activations in each dimension
are correlated with language-level concepts
contained by prompts~\cite{bills2023}.
Even in the linear probe example above,
where features ostensibly correspond to
physical, numeric concepts like time,
those notions are rooted in a model's
general language-based training data.

Because the starting point for a mechanistic probe
is language,
a probe is sensitive to the linguistic framing used.
For example,
consider research into how LLMs
internally represent the state of an Othello game.
Initial work observed that linear probes
were less accurate than nonlinear ones
at recovering game state,
suggesting that LLMs built a nonlinear world model~\cite{li2022emergent}.
However, subsequent work~\cite{nanda2023emergent}
found that
linear probes can be just as effective
if they reflect \textit{the LLM's own perspective} on the game---
during a particular turn in the Othello game,
the LLM does not conceive of a cell
as being ``empty,'' ``black,'' or ``white,''
but rather ``empty,'' ``mine,'' or ``theirs''
(where the color associated with ``mine'' and ``theirs''
flips during every turn).

From the security perspective,
the sensitivity of mechanistic probes to
the linguistic biases of the monitor is troubling.
Ideal sandbox policies would be crisply specifiable
using formal semantics for good and bad behavior.

\subsection{The Security Implications of Linguistic Illegibility}

The bulk of an LLM's computation is
linear and nonlinear math
over real-valued activation vectors.
Only during input embedding and output unembedding
does the model explicitly manipulate reified linguistic constructs.
Prior research on feature superpositioning
suggests that a model's internal representation space
contains more usable features than
the size of the token vocabulary space~\cite{bricken2023monosemanticity}.
Thus,
we are confronted with a fundamental mismatch of capacity:
there are more unique features than unique tokens,
meaning that there must be some features which defy
a reasonable single-token description.
If we could automatically enumerate such features,
we could at least empirically observe,
for each one,
which kinds of multi-token linguistic inputs
most strongly trigger the feature.
However,
the number of possible feature directions
in even a moderately-sized LLM
is vast~\cite{elhage2022,gao2024},
thwarting attempts to simply
iterate through the full catalog of directions and
identify the most relevant linguistic concepts
for each one.

The specter of linguistic illegibility
therefore seems to be inescapable,
at least with our current understanding
of internal model dynamics.
Most of the LLM's vector math
does not directly manipulate language at all
and thus defies direct linguistic analysis;
furthermore,
the model's internal concept space is so vast that
an exhaustive study of the
\textit{empirical} correlation between all features
and all relevant language constructs
is intractable.
The implication for model sandboxing
is not that language-based oversight
is worthless.
However, such oversight \textit{must} be layered atop
isolation mechanisms which do not inherit
the linguistic illegibility of the model itself.
Recent sandbox escapes by frontier models~\cite{anthropic-escape2,anthropic-escape1,oa-escape1}
further suggest that
commodity isolation mechanisms like
containerization are woefully inadequate for
sandboxing highly-capable models.
We propose approaches that we think are more robust
in the next section.

%

%
\section{A Stronger Sandbox}
\label{sec:stronger-sandbox}

\subsection{Threat Model}
\label{sec:threat-model}

We define an \textbf{agent} as
an \textbf{LLM} that runs atop an \textbf{inference engine},
with an \textbf{orchestrator} allowing the inference engine
to interact with the outside world
(e.g., via MCP calls).
In our threat model,
the primary adversary is the agent itself,
so our core security goal is to prevent the agent
from externalizing harmful behavior
that impacts entities outside the agent's \textbf{sandbox}.
For example,
a misaligned model might try to
exfiltrate its weights,
exchange network traffic with non-allowlisted servers, or
modify audit logs associated with its own execution.

Our threat model is
directly motivated by recent sandbox escapes
of models being evaluated:
\begin{smitemize}
  \item \textbf{Example 1:}
        An OpenAI agent escaped a network-isolated sandbox
        by exploiting a zero-day vulnerability in a trusted
        package-manager server that was accessible to the model
        and had open Internet access itself~\cite{oa-escape2,oa-escape1}.
  \item \textbf{Example 2:}
        Anthropic observed that Mythos could
        leverage a multi-step browser exploit to escape from
        the browser's renderer sandbox and then the OS's container
        sandbox~\cite{anthropic-escape1}.
  \item \textbf{Example 3:}
        Anthropic separately observed three cases in which
        models escaped their sandboxes due to configuration errors
        in the sandbox environment (as opposed to
        enforcement failures as the other examples we list above).
        The models then hacked into servers on the open
        Internet~\cite{anthropic-escape2}.
\end{smitemize}
In this section,
we focus on security guarantees that can be provided
even if linguistically-rooted oversight fails,
so we mostly ignore language-based oversight mechanisms.
In practice,
we expect (and encourage!) language-based monitors to form
an inner layer of the sandbox.

\subsection{Architectural Overview}
\label{sec:arch}

\begin{table}
  \caption{The design principles for our proposed sandbox.\vspace{-1mm}}
  \centering
  \begin{tabular}{lcc}
    \toprule
    \textbf{Design principle} & \textbf{Mechanism} & \textbf{Sandbox failure addressed} \\
    \midrule
    \midrule
    Provenance labeling   & \makecell{Taint tracking, information\\ flow control}  & \makecell{Undetected model influence\\ on sensitive state}     \\
    \midrule
    Robust virtualization & \makecell{Hardware-enabled TEEs,\\ verifiable microkernels} & \makecell{Enforcement bypass via\\ privilege escalation}      \\
    \midrule
    Egress mediation      & \makecell{Minimal default-deny\\ proxy w/no agent\\ credentials}       & \makecell{Confused-deputy compromise\\ of trusted infrastructure}  \\
    \midrule
    Behavioral anomaly detection & \makecell{Statistical monitoring of\\ permitted output streams} & \makecell{Information laundering (e.g.,\\ steganography), behavioral phase\\ shifts}      \\
    \midrule
    Configuration attestation   & \makecell{Pre-launch validation\\ by third parties}  & \makecell{Inconsistency between policy\\ goals and policy implementation}     \\
    \bottomrule
  \end{tabular}
  \label{tab:design-principles}
\vspace{-0.1mm}
\end{table}

Our proposed sandboxing architecture
is guided by five principles,
as shown in Table~\ref{tab:design-principles}.
We discuss each principle
(and its consequences for the sandbox architecture) below,
and describe the open research questions.
The first principle (provenance labeling)
has the most subtle design consequences,
so we focus most of our discussion on this principle.

\vspace{-2mm}

\paragraph{Provenance labeling:}
Linguistic illegibility challenges our ability
to understand a model's intent.
But regardless of whether we can understand model intent,
we can still define various system-level interactions
that should be forbidden.
An important kind of unsafe interaction
involves \textit{agentic influence over sensitive bytes
in memory or in persistent storage devices}.
For example,
an agent that can send data
to unauthorized OS-level sinks like
file systems and network sockets
can inappropriately memorialize data
or attack remote endpoints.
As another example,
if an agent can overwrite intra-process state
(e.g., return addresses) that
belong to inference-stack components
like package managers or RAG databases,
the agent can subvert
the control flow of those components~\cite{abadi2009cfi,shacham2007return}.

Our proposed sandbox uses
taint tracking~\cite{enck2014taintdroid,newsome2005dynamic,schwartz2010dynamic}
to taint model outputs
and observe how they influence
other system state.
The goal is to (1) define a set of sinks
which should not receive
model-influenced data, and (2)
define introspection points, akin to LSM hooks~\cite{wright2002linux},
for which security policies can register interest.
An initial implementation of the sandbox
would run on commodity hardware
and define sinks with respect to
OS-level abstractions like
processes, files, and network sockets;
TaintCheck~\cite{newsome2005dynamic} and TaintDroid~\cite{enck2014taintdroid}
are conceptual representatives of the proposed approach.
Ultimately,
we would want hardware-level tainting support~\cite{venkataramani2008flexitaint,wang2023raft}
to efficiently understand how agent output
influences both OS-level sinks
and hardware-level, intra-process state for
components residing outside of the narrow agent sandbox
(but inside the inference-as-a-service deployment).

Note that the last sentence implies that
the full deployment environment
runs atop a taint-tracking framework.
This requirement is non-trivial
but reasonable if one takes the threat of
superintelligent misaligned agents seriously;
being able to identify
unexpected model influence
throughout the inference stack
is critically important.

The taint sinks defined in the classic literature
(e.g., system-call-level abstractions,
intra-process control-flow state like
return addresses and function pointers)
are all relevant in our scenario.
However,
what should be the taint \textit{sources},
and how should \textit{detainting} occur?
A naive approach would not support detainting,
and would use single-bit taint tags
which indicate whether \textit{any} of the model's output tokens
have influenced the tagged state.
The risk of this straightforward approach
is overtainting~\cite{dalton2010tainting,hough2022conflux,slowinska2009pointless}
in which the system monotonically accumulates taint
(such that the taint tags gradually become
less useful to security policies).
To avoid these problems,
we propose \textit{causal tagging of output tokens}
and \textit{auditable detainting}.
We describe these concepts below.

At a high level,
causal tagging allows the sandbox
to have many taint tags.
Some tags are associated with actors
(e.g., users who supply prompts,
or different instances of deployed agents);
other tags are associated with specific resources
like files in a RAG database or
outputs from a tool invocation.
An output token from an agent
receives the union of tags
from the relevant actors and resources
that the agent's model consulted
to generate the token.
For example,
some output tokens might be influenced
(or not influenced) by
the system prompt, the user prompt,
the output of a specific tool call,
or the contents of a document fetched by RAG.
Other tokens might be primarily influenced by
a model's intrinsic biases towards
generating certain output patterns.
By tagging each output token
with the information used by the model to derive that token,
we enable rich sandboxing policies, e.g.,
\begin{smitemize}
  \item \textbf{No agentic output that is tainted by documents from
        internal RAG databases should leave the internal network:}
        This policy thwarts agent-driven exfiltration of
        sensitive corporate data.
  \item \textbf{An output sink owned by entity $e$ should only receive
        agentic output that is tainted by inputs from tools that
        are also owned by $e$:} For example, if each tool is tagged
        with the identity of a particular customer, and each output 
        sink is also tagged with the identity of the owning customer,
        then no customer can receive information derived
        from data belonging to a different customer.
  \item \textbf{A privileged deputy, when provided with tainted inputs,
        can only perform actions that are allowed by the capabilities of the taint sources:} For example, a tool gateway that
        internally manages upstream API keys should only use
        a key in response to inputs that were generated by
        authorized users of that key.
\end{smitemize}
Besides these contextual policies,
the sandbox can also support
more straightforward categorical policies
(e.g., ``no model-tainted bytes should \textit{ever}
reach a particular sink'').
Attempted policy violations should be logged at a minimum,
and (depending on the seriousness of the violation)
should be grounds for halting the agent
or buffering its outputs for asynchronous analysis
before the outputs are
detainted, logged, and externalized.

The best algorithm for
assigning tags to output tokens
is an open research question.
We are exploring three different approaches,
all of which assume that
input tokens are pretagged
and that tags are floating point numbers
between 0.0 and 1.0 representing
the strength of a taint source's influence.
\begin{smitemize}
  \item With \textit{attention-based output tagging},
        the inference engine looks at the transformer's
        attention patterns over input tokens.
        During a forward pass, tags from strongly-attended 
        tokens propagate their taint more aggressively to 
        the new output token.
  \item Alternatively, the inference engine might  
        calculate the Jacobian or the integrated 
        gradient~\cite{sundararajan2017} of the LLM's output logit with respect to the input tokens,
        and scale propagation of \textit{gradient tags}
        with respect to the magnitude of an input token's
        gradient.
  \item With \textit{counterfactual-based output tagging},
        the impact of a particular input token on an output
        token (or full response) is determined via an
        ablation experiment in which the
        model is rolled back to an earlier snapshot,
        the token in question is replaced by a filler token,
        and the model's forward passes are re-executed to
        generate a new response~\cite{neurotaint2026}.
        If the original response and the
        counterfactual one differ significantly (e.g.,
        output different tool calls or contain significantly
        divergent linguistic content), the ablated input token
        propagates its tag to the agent's output.
\end{smitemize}
Attention-based tagging can run
inline with a normal forward pass,
but will not always provide true causal attribution~\cite{jain2019attention,pruthi2020deceive}.
Counterfactual-based tagging does,
but requires additional forward passes
and thus is better suited for
batch validation of a buffered multi-token response.
Also note that
attention-based and gradient-based tag propagation
require introspection on intra-GPU data flows
(e.g., using instrumented GPU kernels~\cite{hayes2017,nvbit}
or native GPU support for taint tracking~\cite{hu2021}).


\longpage
\vspace{-2mm}
\paragraph{Robust virtualization:}
Taint propagation, sink monitoring, and policy enforcement
must execute in a strictly more-privileged layer
than the agent's own.
For example,
the agent could run atop sel4~\cite{klein2009sel4},
a verified microkernel with provable guarantees
of isolation and noninterference.
Alternatively (or additionally),
agents could be executed within
hardware-rooted virtualization layers
with information flow tracking and policy decisions
implemented by code that hardware prevents the agent from modifying.
In either case,
the goal is to ensure that an agent which
has achieved full privilege escalation within its guest
cannot disrupt sandbox machinery.

\vspace{-1mm}
\paragraph{Egress mediation:}
All of an agent's outbound network activity
is forced through a minimal network proxy.
The proxy rejects all connections by default,
only allowing those enabled by a manifest
signed by the first-party inference operator
and a third-party validator
(see our discussion of configuration attestation below).
The proxy only acts as a firewall,
and contains no additional functionality
like package management;
the goal is to reduce the proxy's threat surface.

\vspace{-1mm}
\paragraph{Behavioral anomaly detection:}
Taint tracking cannot detect implicit flows~\cite{hough2022conflux}
in which information is externalized not through tainted bytes,
but through a choice of which untainted bytes to externalize;
for example,
agents may try to launder information through
token-distribution biasing~\cite{silva2024}
or zero-width Unicode formatting~\cite{arxiv260308104},
or through selective invocation of seemingly innocuous tool.
An agent that is not trying to leak information,
but simply disrupt internal or remote systems,
may also exhibit phase changes in behavior, e.g.,
suddenly issuing a large number of tool calls
during the reconnaissance phase of remote server exploitation.
A robust sandbox
must therefore perform statistical analysis of
the content in ostensibly low-risk output channels,
and rate-limit, pause, or subject to greater scrutiny
models that are deviating from expected behavioral baselines.

\vspace{-1mm}
\longpage
\paragraph{Configuration attestation:}
Before a model is deployed
(whether for training, testing, or real inference serving),
the sandbox configuration must be independently verified
by an organizationally and infrastructurally separate entity.
During the verification period,
the verifier has complete access to
the computational infrastructure of the first-party agent vendor,
allowing the verifier to (for example)
run test agents directly on the first-party infrastructure
and perform red-teaming on that infrastructure.
Once verification is done,
the sandboxing hardware and software
expect the sandboxing configuration to be co-signed by
both the independent third-party
and the agent's first-party vendor.
Any changes to that configuration
must also be co-signed.

\subsection{Effectiveness and Limitations}
\label{sec:limitations}

Returning to Section~\ref{sec:threat-model}'s examples
of real-life sandbox escapes,
we see that our proposed sandbox infrastructure
would mitigate or outright prevent those escapes:
\begin{smitemize}
  \item \textbf{Example 1:} Even if the OpenAI agent was able to
        subvert the package manager, the manager would not
        have direct access to the public Internet. Instead, the
        manager would connect to the minimal network proxy; the uncompromised proxy (or if that is
        compromised, the uncompromized
        taint-tracking system) would detect that model-influenced
        bytes were trying to leave the internal network, and
        either unconditionally raise an alarm or verify that
        the model-influenced bytes are acceptable (e.g., because they
        only represent the name of the package that the model
        wants to fetch, in which case the sandbox would write
        the detainting decision to a tamper-proof log).
        Behavioral analysis would also flag that abnormally
        large number of tool calls that misaligned agents
        issued while probing external systems.
  \item \textbf{Example 2:} Mythos could still escape from the
        browser-level sandbox, but attempts to escape from the virtualization-level sandbox would fail (e.g., due to
        the formal security guarantees of sel4, or the inability
        of Mythos to corrupt isolation mechanisms protected
        by the hardware itself).
  \item \textbf{Example 3:} Anthropic's intended sandbox configuration
        was correct; the deployed configuration was not. Configuration
        attestation would detect the mismatch and force Anthropic to
        fix the problem before deploying the models atop the (now
        properly-configured) sandbox. Taint tracking would allow
        the sandbox to prevent model-influenced bytes from traversing
        sinks connected to the public Internet.
\end{smitemize}
In summary,
we think that the proposed sandbox would
materially increase the difficulty
of subverting isolation boundaries.

That being said,
building the proposed sandbox is not trivial.
For example,
configuration attestation will require
careful technical and legal design
to ensure the integrity of first-party infrastructure,
respect intellectual property,
and establish clear liability regimes.
The anomaly detection layer is fundamentally statistical,
meaning that agents can try to evade it by
encoding problematic behavior in ways that
nonetheless satisfy the sandbox's
statistical understanding of normal behavior.
Guarantees provided by formally verified hypervisors like sel4
are premised on the correctness of the proofs;
in turn,
proof correctness relies on the trustworthiness
of the compiler and the hardware which executes the compiled code.
Hardware-level side channels~\cite{kocher2019spectre,lipp2018meltdown}
are also a constant concern,
although they can be mitigated
by reducing or eliminating
microarchitectural co-tenancy~\cite{hofmann2024,ramkrishnan2026}.

Another important caveat is that
taint tracking can monitor explicit flows
but not implicit ones (\S\ref{sec:arch}).
The behavioral anomaly detector
can identify some implicit flows,
but only in a probabilistic sense.
If we replaced taint tracking with
information flow control~\cite{denning1976lattice,flume,histar,asbestos},
the sandbox could detect such implicit flows;
however, monitored applications would have to be
(1) written in an IFC language like Jif~\cite{jif},
or (2) decomposed into privileged-separated components
involving non-secret workers which send requests
to secret-handling workers, who then compute over
sensitive data and give the results to
a trusted declassifier for
subsequent return to non-secret workers~\cite{flume}.
In the short term,
rewriting the inference stack to use IFC
seems more daunting than porting legacy applications
to taint tracking frameworks,
but ultimately,
an IFC inference stack could enforce policies
with much lower runtime overhead---taint tracking
essentially requires every assignment operation
in a program to be instrumented~\cite{taintrabbit},
whereas IFC allows inappropriate data transfers
to be caught statically or
with simple runtime checks enforceable by the kernel
during system calls.
\section{Related Work}
\label{sec:related}

\vspace{-2mm}
\paragraph{Dynamic taint tracking:}
This technique has a rich history in the literature.
For example,
see prior work on
whole-system taint tracking for exploit detection~\cite{taintbochs},
control-flow-aware propagation policies
to combat overtainting~\cite{hough2022conflux},
and hardware-assisted taint propagation
to decrease runtime overhead~\cite{venkataramani2008flexitaint,wang2023raft}.
Classic taint tracking
examines data flows in programs with
imperative def-use updates of discrete program variables;
in LLMs,
program state is more diffuse,
consisting of high-dimensional vectors whose
intrinsic meaning is less obvious and that are updated by
continuously-valued mathematical operations.

\vspace{-2mm}
\paragraph{Information flow control for LLM agents:}
Several recent systems leverage IFC
to improve LLM security.
FIDES~\cite{fides2025} attaches confidentiality and integrity labels
to orchestrator inputs and outputs
like tool arguments and results,
preventing untrusted inputs from causing prompt injection attacks.
FIDES also prevents the explicit leaking of sensitive data,
but does not prevent implicit leaks because
agents are allowed to make control-flow decisions on secret data.
NeuroTaint~\cite{neurotaint2026} detects implicit flows involving secret data
using a counterfactual probe that resembles the one
discussed in Section~\ref{sec:arch}.
CaMeL~\cite{CaMeL} handles implicit flows
by forcing the model to emit a concrete action plan
in the form of Python code;
CaMeL executes the plan in a modified Python interpreter
that enforces IFC policies.
FIDES, NeuroTaint, and CaMeL focus on
how untrusted data might trick
a trusted model into behaving inappropriately;
in contrast, our focus is on sandboxing a full inference deployment
where the model itself might be intrinsically misaligned.

\vspace{-2mm}
\paragraph{AI control:} This line of work
assumes that a model can be misaligned,
and investigates oversight protocols
in which the output of the untrusted model
is reviewed by a weaker but trusted model
(or by a strong, untrusted model deployed atop
an evaluation framework which discourages collusion
between the two models)~\cite{aicontrol}.
These approaches fundamentally require
analysis of a model's natural-language content,
and thus are exposed to validity threats
posed by linguistic illegibility.

\section{Conclusion}

Linguistic illegibility is the semantic gap
between what a model's language output reveals
and what a model's internal computation actually does.
We argue that linguistic illegibility
is always possible with models that
do not intrinsically compute over reified language concepts,
but instead compute over activation spaces that are
lossily initialized from and converted to natural language
at the bookends of model computation.
If linguistic illegibility is inevitable in at least some situations,
then sandboxing methods that are rooted
in natural-language analysis
(e.g., chain-of-thought analysis,
constitutional self-critique,
and mechanistic probing)
\textit{must} be supplemented by systems-level mechanisms 
whose policies are not rooted in natural-language analysis.
Furthermore,
if the threat model includes the possibility
that models are competitive with or surpass human capabilities,
the systems-level sandbox must be
relentlessly paranoid and incredibly robust:
paranoid because the sandbox must track
a wide variety of model behaviors,
and robust because the sandbox must not be subvertible
by a highly-capable model.
We think that our proposed design principles
are a good starting point for
what a paranoid and robust sandbox
should look like.
\section*{Acknowledgments}
This work was partially supported by funding from
Georgetown University's Center for Security and
Advanced Technology, and from Coefficient Giving
(formerly known as Open Philanthropy).
\bibliographystyle{plain} 
\bibliography{refs}


\end{document}